\documentclass{article}

\usepackage[utf8]{inputenc}
\usepackage{amssymb}
\usepackage{amsmath}
\usepackage[english]{babel}
\usepackage{graphicx}
\usepackage{hyperref}
\usepackage{multirow}
\usepackage[T1]{fontenc}
\usepackage{url}
\usepackage{graphicx}
\usepackage{booktabs}
\usepackage{nicefrac}
\usepackage{microtype}
\usepackage[table,x11names]{xcolor}
\usepackage{authblk}

\title{Pre-Training Transformers for Domain Adaptation}

\def\correspondingauthor{\footnote{Corresponding author: burhan@shirleyrobotics.com}}
\author[1]{Burhan Ul Tayyab \correspondingauthor{}}
\author[1]{Nicholas Chua}
\affil{Shirley Robotics}

\date{}

\begin{document}
\maketitle

\section*{Abstract}
The Visual Domain Adaptation Challenge 2021 called for unsupervised domain adaptation methods that could improve the performance of models by transferring the knowledge obtained from source datasets to out-of-distribution target datasets. In this paper, we utilize BeiT \cite{DBLP:journals/corr/abs-2106-08254} and demonstrate its capability of capturing key attributes from source datasets and apply it to target datasets in a semi-supervised manner. Our method was able to outperform current state-of-the-art (SoTA) techniques and was able to achieve 1st place on the ViSDA Domain Adaptation Challenge with ACC of 56.29\% and AUROC of 69.79\%.

\section{Introduction}
\subsection{Visual Domain Adaptation}
Traditional deep learning methods work really well in a constrained environment where the target dataset is close to the source dataset on which it is trained on. However, as demonstrated by \cite{Barbu2019ObjectNetAL}, any shift in attributes (viewpoints, lightning conditions, orientations etc) and/or shift in label classes (where the target set varies/has new classes which aren’t present in source dataset) would could cause the model to perform poorly and the accuracy to drop significantly. This could lead to a lot of problems in real-life scenarios if didn’t taken into account. To solve this problem, we utilize BeiT \cite{DBLP:journals/corr/abs-2106-08254}, and demonstrate that it could self-learn various attributes by itself and can be adapted on new target datasets.

\subsection{Related Work}
Following AlexNet\cite{NIPS2012_c399862d}, convolutional neural networks (CNNs) have become standard for image classification tasks. Various models based on CNNs\cite{he2016deep}\cite{xie2017aggregated}\cite{tan2019efficientnet} have been introduced that achieve a significant increase in accuracy on various datasets\cite{ILSVRC15}. However, these models fail to perform in conditions where there is a big input distribution shift between training and testing dataset and/or has label set variance\cite{Barbu2019ObjectNetAL}. Transformers such as BiT\cite{kolesnikov2020big}, ViT\cite{dosovitskiy2020image} have demonstrated significant improvement over CNNs in terms of accuracy at image classification tasks, however they require huge amounts of data to train. Meanwhile\cite{NIPS1999_5a142a55} utilizes the idea of using a single fixation of parse trees for image classification and attribute learning and\cite{DBLP:journals/corr/abs-1710-09829} improves that by using dynamic routing and fixed vector representation for image classification, however both of the ideas don’t work significantly well on large datasets because of its nonlinearity which results an increase in training complexity.

Domain Adaptation refers to fitting a model that has been trained on a particular source dataset on an out-of-source novel target distribution, which is not part of the training set. Closed-Set Domain Adaptation\cite{long2015learning}\cite{tzeng2017adversarial} methods, where the source and the target domain completely share the class of their samples, work extensively well and have low input distribution shifts, however they fail to work in open-set environments because of unknown target input samples. Self-supervised learning methods could be used to solve these issues by either distillation\cite{belal2021knowledge} or contrastive learning\cite{kang2019contrastive}, however these methods have significant drawbacks\cite{piva2021exploiting}\cite{tang2020unsupervised}.

\section{Proposed Approach}
In this section, we will show the proposed method in detail. Figure-1 gives an overview of the method utilized by us. 

\subsection{Model: BeiT}
BERT\cite{devlin2019bert}, along with its Masked Language Modelling (MLM) module has performed wonders in the Natural Language Processing (NLP) domain. Inspired by BERT, we utilize BeiT-B\cite{DBLP:journals/corr/abs-2106-08254} for performing universal domain adaptation. The input image is preprocessed and converted into patches, while is also simultaneously tokensized by DALL-E\cite{ramesh2021zero}. The patches are then masked randomly and fed to the BeiT-B Encoder which outputs hidden embeddings which are reconstructed by Masked Image Modelling Head by using input tokens. Since the idea is to just train the method on ImageNet-1k and test it on a related but un-constrained dataset with open world settings, we chose this method because it resembles masked language modelling in BERT. The whole system is pre-trained on ImageNet-1k, where the Masked Image Modelling module is able to reconstruct the corrupted patch via self-supervised self-attention and thus is able to recognize and separate the image without using any labels. After pretraining, a classification head is attached to the model where the pretrained model is fine-tuned to perform image classification.

\begin {figure}
\begin {center}
\includegraphics[width=0.95 \textwidth] {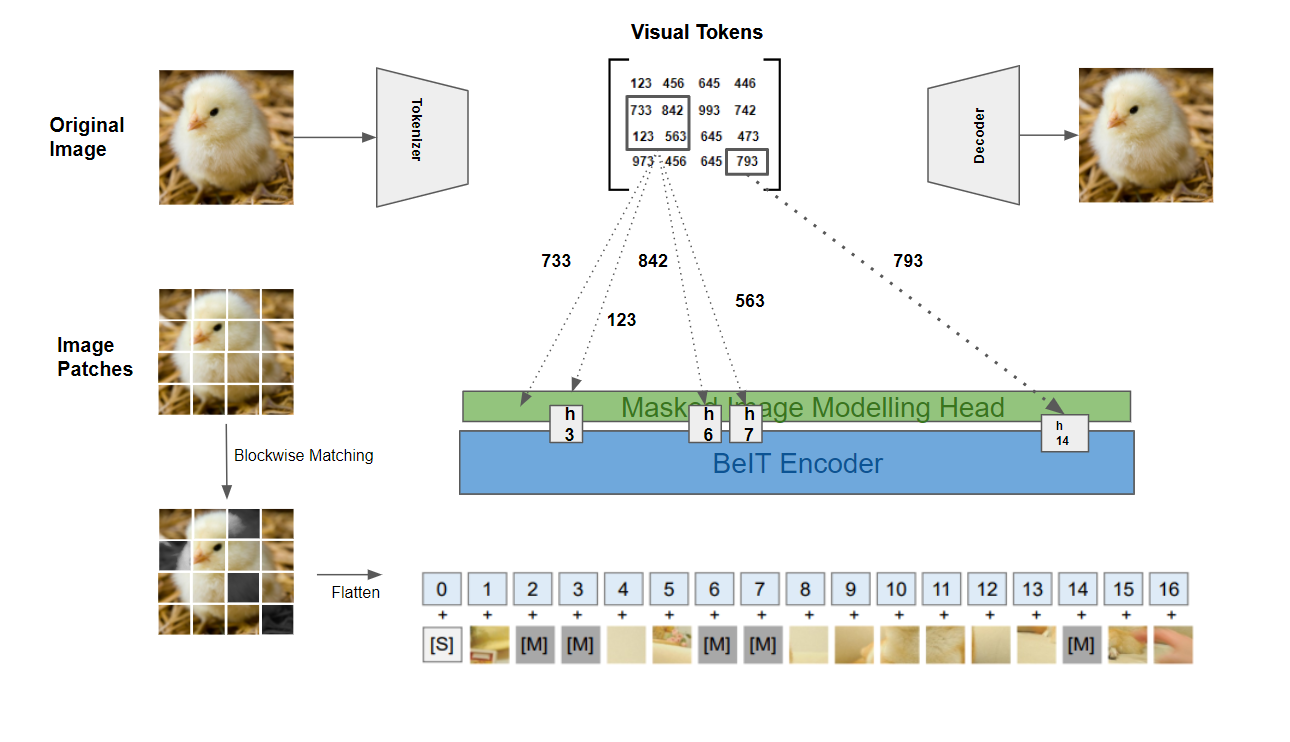}
\caption{BeiT Architecture}
\end {center}
\end {figure}

\subsection{Preprocessing}
For pretraining, the dataset was resized into 224 x 224 and preprocessed with random cropping, color jittering, and horizontal flipping. The input image is then split into 14 x 14 images patches and masked via block-wise masking\cite{DBLP:journals/corr/abs-2106-08254}. Simultaneously, the input image is also tokenized via \cite{ramesh2021zero} which would later be utilized for performing visual patch reconstruction by Masked Image Modelling head.

\subsection{Methods}
\subsubsection{L-Layer ViT (BeiT)}
We denote a dataset of images as $X=\{x_1,...,x_N\}$ where $N$ represents the total number of images and $\forall i \in \{1,...,N\},\ x_i \in \mathbb{R}^{H\times W \times C}$. For  image $x_i$, we will splice it into d number of image patches which is represented as $x_i^p \in \{x^p_{(i,1)},...,x^p_{(i,d)}\}$ where each patch is denoted by $x^p_{(i,k)} \in \mathbb{R}^{P\times P\times C}$, the number of patches are $d=\frac{HW}{P^2}$, and $P^2$ is the area of the patch. After slicing, the patches are masked randomly via blockwise masking algorithm \cite{DBLP:journals/corr/abs-2106-08254}. The patches are then flattened to form a matrix, $v_i^p \in \mathbb{R}^{(P^2C)\times d}$. The embedding matrix $E \in \mathbb{R}^{D\times (P^2C)}$ and position embedding matrix $E_{pos} \in \mathbb{R}^{d\times D}$ are linearly are linearly embedded to the image patch which is fed to the through the $L$ Layers of the transformers to produce a set encoding vectors  $H_L$.

\subsubsection{Masked Image Modelling Head}
The dataset of images $X=\{x_1,...,x_N\}$ are also fed into image tokenizer\cite{ramesh2021zero} which converts them into $Z=\{z_1,...,z_N\}\in {V}^{H \times W}$ tokens, where $W$ denotes the width of the image, and $H$ denotes the height, whereas $V$ is the vocabulary which contains $V=\{v_1,...,v_N\}$  discrete token indices.

\subsubsection{Pretraining}
We chose to utilize BeIT for domain adaptation because of its semi-supervised self-attention mechanism. The model consists of a 12-layer transformer with hidden size of 768 and 12 attention heads. The input image is resized into 224 x 224 resolution, anc converted into 14 x 14 image patches where each patch is of size 16 x 16. Simultaneously, the input image is also tokenized into 14 x 14 semi-tokens. 40\% of the patches are randomly masked via blockwise masking algorithm and masked regions are attached with learnable embeddings. The patches are then fed into image transformer (encoder), where it produces the output, which is fed into masked image modelling head, which takes the input tokens and patches and tries to reconstruct the corrupted masked patch. The pre-training objective is to
minimize the loss between original token and reconstructed token derived from
the patched image. The loss function is be denoted by

\begin {equation}
	\sum_{(x_i, \tilde{x_i} \in D)} (\mathbb{E}_{z_i \sim q_\phi (z | x_i)} [\log p_\psi (x_i | z_i)]  +  log p_\theta (\tilde{z_i} | \tilde{x_i})   ) 
	\label{sum}
\end {equation}

where  $\mathbb{E}_{z_i \sim q_\phi (z | x_i)} [\log p_\psi (x_i | z_i)]$ denotes visual token reconstruction loss and $log p_\theta (\tilde{z_i} | \tilde{x_i})$ represents masked image modelling loss. Here $x$ denotes the original image $\tilde{x_i}$ represents corrupted masked image and ${z}$ denotes tokens obtained from the tokenizer; $p$ is the pretraining objective which is maximize the log-likelihood of the corrupt visual tokens given a corrupted image and $D$ represents the training dataset, $q_\phi (z | x_i)$ denotes the image tokenizer, $p_\psi (x_i | z_i)$ is the function that decodes original image from visual tokens and $ p_\theta (\tilde{z_i} | \tilde{x_i})$ recover the visual image patches from corrupted patches.

\subsubsection{Fine-tuning}
After successful pretraining, a classification head (fully-connected network) is attached to the model, after which the model is fine-tuned on ImageNet-1k for another 500 epochs. For the first 400 epochs, the image size is the same as pre-training input (224 x 224), however, for the last 100 epochs, the images are reshaped and fed as having 384 x 384 resolution. 

\section{Experiment}
\subsection{Datasets}
Due to the model size / training constraint in ViSDA 2021 Challenge \cite{bashkirova2021visda2021}, the training dataset provided was ImageNet-1k \cite{ILSVRC15} which contains 1.4M images and 1000 classes. There were also 3 development datasets provided, as shown in Table 1. which were

\begin{enumerate}
    \item ObjectNet \cite{Barbu2019ObjectNetAL} consisting of 50,000 images with 313 classes where only 113 classes are the same as the source dataset and the images are generally more difficult to classify due to the large differences in poses and backgrounds between each image of the same class.

    \item ImageNet-R \cite{hendrycks2021faces} which contains 30,000 images with 200 classes from the source dataset with varying visual styles and textures.

    \item ImageNet-C \cite{hendrycks2019benchmarking} which is similar to ImageNet however the images are corrupted.
\end{enumerate}

The development datasets weren’t allowed to be used for training purposes. However other than that, they could be utilized for model development by tuning it’s hyperparameters.

\begin{table}[h]
\begin{center}
    \begin{tabular}{|c|c|c|p{2cm}|}
    \hline
    Dataset & Number of Images & Number of Classes & Note*\\
    \hline
    ImageNet(source) & 1.4M & 1000 & \\
    \hline
    ObjectNet & 50,000 & 313 & Only 113 classes are the same as the source\\
    \hline
    ImageNet-R & 30,000 & 200 & Different texture/style\\
    \hline
    ImageNet-C & 1.4M & 1000 & Corrupted\\
    \hline
\end{tabular}
    \caption{Dataset Description}
\end{center}
\end{table}

\subsection{Training}
For the experiment, only ImageNet-1k for training purposes. Even though ImageNet-C, ImageNet-R and ObjectNet were also allowed to be used for tuning hyperparameters from the pretrained source model, we in fact, didn’t use it. Since our model was able to perform visual reconstruction on target visual token via self-supervised attention mechanism, we believed that it can also be able to learn various attributes from various images and that information could be cross-applied to different target datasets \cite{devlin2019bert}. Moreover, we discovered that the model could also auto-seperate objects and classes without labelling. We basically created a random and corrupted mesh of ImageNet-1k images with random noise and jittering to create a new class in existing dataset to tackle out-of-distribution classes in the target dataset. We do believe, however that pretraining self-supervised models on ObjectNet, ImageNet-C and ImageNet-R will increase the accuracy of the model.

\subsection{Implementation Details}
The model was pre-trained on ImageNet-1k. The augmentation used consisted of color-jittering, horizontal flipping and random resized cropping. We ran the pretraining for 500k steps with 1k batch size. The learning rate of 1.5e-3 and cosine learning weight decay of 0.05 is used. Adam optimizer with B1 = 0.9 and B2 = 0.999 is used. The training of 500k steps  (1600 epochs) takes about 12 days using Nvidia Tesla V100 32GB GPU cards.

\begin{table}[h]
\begin{center}
\begin{tabular}{|c|c|c|}
    \hline
    Methods & Parameters & ACC \\
    \hline
    ImageNet-1K(pretraining) & 86M & 82.4\% \\
    \hline
    ImageNet-1K(pretraining + fine-tuning) & 86M & 83.0\% \\
    \hline
\end{tabular}
    \caption{Ablation Studies}
\end{center}
\end{table}
The above table shows the top-1 accuracy trained on ImageNet-1k. The model achieves very high accuracy compared to various state of the art techniques \cite{dosovitskiy2021image, kolesnikov2020big}, which generally require larger datasets \cite{schuhmann_2021, sun2017revisiting}.

\subsection{Results: Performance on ViSDA 2021 Challenge}
Our model ranked 1st place on the VisDA 2021 leaderboard,outperforming the second place by 15.04\% on source-only accuracy and 7.73\% on adapted model accuracy. To show that our model can outperform any existing domain adaptation techniques, we didn’t use any development sets at all. Moreover, Table 3. shows that our adapted AUC is slightly lower than some of the other entries, that is truly understandable as the model was actually picking attributes from the source and applying them onto target sets rather than performing a max-mean discrepancies models which tries to reduce distance between source and target features. This clearly shows that our model is capable of picking attributes from the source and it can link those attributes from source-to-target in a completely self-supervised manner.

\begin{table}[ht]
\begin{center}
    \resizebox{\columnwidth}{!}{
        \begin{tabular}{|c|p{2cm}|p{2cm}|p{2cm}|p{2cm}|}
    \hline
        Methods & {\small ACC(Adapted Model)} & {\small AUC(Adapted Model)} & {\small ACC(Source model)} & {\small AUROC(Source model)} \\
    \hline
      \rowcolor{lightgray} {\small \textbf{babychick(ours)}} & 56.29 & 69.79 & 56.29 & 69.79 \\
    \hline
        {\small liaohaojin} & 48.56 & 70.72 & 41.25 & 64.48 \\
   \hline
        {\small chamorajg} & 48.49 & 76.86 & 0.07 & 50.00 \\
   \hline
        {\small DXM-DI-AI-CV-TEAM} & 48.60 & 68.29 & 25.70 & 62.43 \\
   \hline
        {\small fomenxiaoseng} & 45.23 & 78.76 & 40.22 & 60.43 \\
   \hline
\end{tabular}
}
    \caption{Test Source results}
\end{center}
\end{table}
\section{Future work}
In this work, we’ve shown that pretraining transformers can successfully be applied for performing Domain Adaptation. In future, we would like to extend this technique to domain adaptation in multi-attribute object detection for successfully transferring the attributes from source to universal target dataset via textual embeddings

\section{Conclusion}
In this paper, we apply pre-train and finetune BeiT on ImageNet-1K and demonstrate that it is able to outperform current state-of-the-art domain adaptation techniques.

\bibliographystyle{unsrt}
\bibliography{NeurIpsVISDA2021}
\end{document}